\documentclass[letterpaper, 10 pt, conference]{ieeeconf}  %

\IEEEoverridecommandlockouts                              %

\usepackage{booktabs}   
\usepackage{multirow}   
\usepackage{tabularx}   

\usepackage{cite}

\usepackage[utf8]{inputenc}
\usepackage[T1]{fontenc}
\usepackage{graphicx}

\usepackage{color}
\usepackage[table]{xcolor}
\usepackage{colortbl}
\definecolor{mygray}{gray}{.9}
\definecolor{ggray}{RGB}{127,127,127}
\definecolor{mygreen}{RGB}{93,174,86}
\definecolor{macaronBrown}{RGB}{205,170,125}

\usepackage{amsmath}
\usepackage{amssymb}
\usepackage{makecell}
\usepackage{float}
\DeclareMathOperator*{\argmin}{arg\,min} 

\makeatletter
\let\NAT@parse\undefined
\makeatother
\usepackage[colorlinks]{hyperref}
\hypersetup{colorlinks=true,linkcolor=red,citecolor=blue,urlcolor=magenta}

\title{\LARGE \bf
CoBEVMoE: Heterogeneity-aware Feature Fusion with Dynamic Mixture-of-Experts for Collaborative Perception
}

\author{Lingzhao Kong$^{1}$, Jiacheng Lin$^{1}$, Siyu Li$^{2,3}$, Kai Luo$^{2,3}$, Zhiyong Li$^{1,2,3}$, and Kailun Yang$^{2,3}$
\thanks{This work was supported in part by the National Natural Science Foundation of China (Grant No. U21A20518, No. 61976086, and No. 62473139), in part by the Hunan Provincial Research and Development Project (Grant No. 2025QK3019), and in part by the State Key Laboratory of Autonomous Intelligent Unmanned Systems (the opening project number ZZKF2025-2-10). 
\textit{(Corresponding authors: Zhiyong Li and Kailun Yang.)}}%
\thanks{$^{1}$L. Kong, J. Lin, and Z. Li are with the School of Computer Science and Electronic Engineering, Hunan University, Changsha 410082, China.}%
\thanks{$^{2}$S. Li, K. Luo, Z. Li, and K. Yang are with the School of Artificial Intelligence and Robotics, Hunan University, Changsha 410012, China (email: zhiyong.li@hnu.edu.cn; kailun.yang@hnu.edu.cn).}%
\thanks{$^{3}$S. Li, K. Luo, Z. Li, and K. Yang are also with the National Engineering Research Center of Robot Visual Perception and Control Technology, Hunan University, Changsha 410082, China.}
}

\begin{document}

\maketitle

\begin{abstract}

Collaborative perception aims to extend sensing coverage and improve perception accuracy by sharing information among multiple agents. However, due to differences in viewpoints and spatial positions, agents often acquire heterogeneous observations. Existing intermediate fusion methods primarily focus on aligning similar features, often overlooking the perceptual diversity among agents. To address this limitation, we propose \textbf{CoBEVMoE}, a novel collaborative perception framework that operates in the Bird's Eye View (BEV) space and incorporates a Dynamic Mixture-of-Experts (DMoE) architecture. In DMoE, each expert is dynamically generated based on the input features of a specific agent, enabling it to extract distinctive and reliable cues while attending to shared semantics. This design allows the fusion process to explicitly model both feature similarity and heterogeneity across agents. Furthermore, we introduce a Dynamic Expert Metric Loss (DEML) to enhance inter-expert diversity and improve the discriminability of the fused representation. Extensive experiments on the OPV2V and DAIR-V2X-C datasets demonstrate that CoBEVMoE achieves state-of-the-art performance. Specifically, it improves the IoU for Camera-based BEV segmentation by ${+}1.5\%$ on OPV2V and the AP@0.5 for LiDAR-based 3D object detection by ${+}3.0\%$ on DAIR-V2X-C, verifying the effectiveness of expert-based heterogeneous feature modeling in multi-agent collaborative perception. The source code will be made publicly available at \url{https://github.com/godk0509/CoBEVMoE}.

\end{abstract}

\section{Introduction}

The perception module serves as the cornerstone of safe autonomous driving operations, providing reliable environmental priors for downstream tasks such as path planning~\cite{9266268}.
Due to limitations in sensor coverage and viewpoint constraints inherent in single-agent perception, multi-agent collaborative perception has emerged as a key paradigm for enhancing the environmental awareness of autonomous driving~\cite{10517450, liu2023, Li_2023_ICCV,wang2023core,chen2023transiff}. 
Multiple intelligent agents, such as vehicles, form collaborative perception systems by sharing sensory information, enabling safe operation beyond visual range.

Due to the complex kinematic interactions among multiple vehicles, the effective fusion of perceptual information from diverse agents constitutes a core technical challenge in collaborative perception systems.
Bird's-Eye View (BEV) represents 3D environmental information in a 2D planar space~\cite{Zhou_2022_CVPR}. The BEV spaces of different vehicles can be aligned solely based on their relative poses, providing an efficient unified spatial framework for collaborative perception systems.
V2VNet~\cite{wang2020v2vnet} introduced spatial transformation to aggregate BEV features from multiple agents. 
CoBEVT~\cite{xu2022cobevt} utilized a Swin Transformer-based fusion backbone with alternating window-grid attention to improve the spatial granularity of multi-agent BEV feature fusion. 
V2X-ViT~\cite{xu2022v2xvit} further incorporates Transformer encoders to model long-range inter-agent dependencies. 
Although these methods significantly enhance collaborative perception performance, they primarily focus on shared semantic features among agents while overlooking their individualized characteristics.

\begin{figure}
    \centering
    \includegraphics[width=1.0\linewidth]{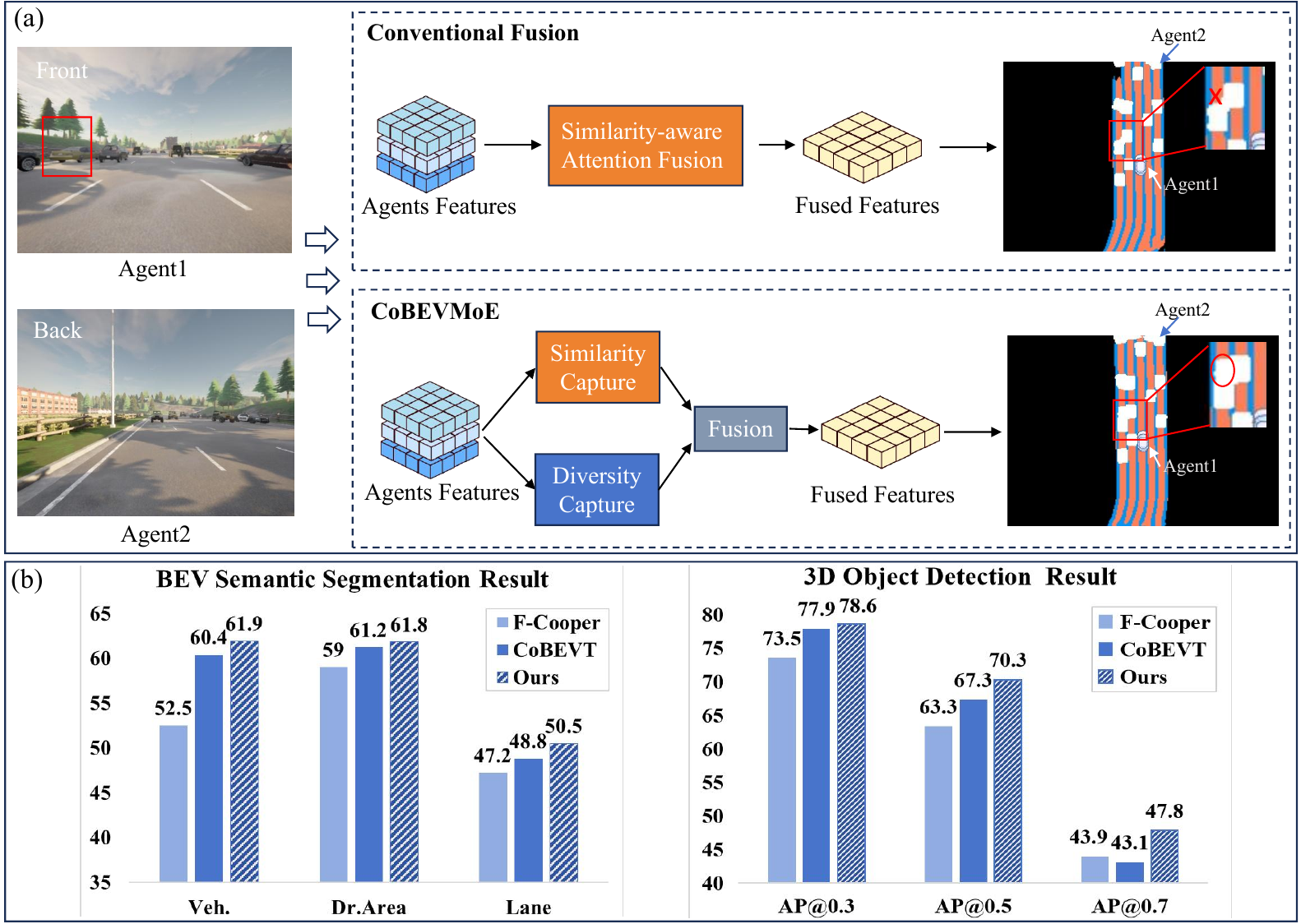}
    \vskip-1ex
    \caption{(a) Conventional attention-based fusion suppresses agent-specific cues, causing missed perception (red box). Our DMoE-based fusion preserves heterogeneous features and correctly segments the target. (b) Quantitative comparison on OPV2V BEV semantic segmentation and DAIR-V2X-C 3D object detection, showing consistent IoU and AP improvements.}
    \label{fig:motivation}
    \vskip-3ex
\end{figure}

Agents with significant positional and orientational disparities often observe unique or complementary features that may be occluded or outside the field of view of other agents.
These viewpoint-specific cues can be highly informative for improving perception, especially under partial occlusions or sensor failures. 
As illustrated in Figure~\ref{fig:motivation}, when a target is observed by one agent but neglected by another, current similarity-guided fusion methods fail to detect this target.
Motivated by this, a successful multi-agent collaboration method should account for both feature commonality and heterogeneity, allowing agents to contribute uniquely based on their observations.

In this work, we propose \textbf{CoBEVMoE}, a novel collaborative perception framework designed to explicitly model both the similarity and diversity of multi-agent observations in the BEV space.
The similar features across different agents can be effectively integrated using well-established attention mechanisms. Simultaneously, this paper aims to achieve parallel learning of differentiated features across various agents using independent modules. Interestingly, this concept is closely aligned with the Mixture of Experts (MoE) paradigm~\cite{shazeer2017outrageously} where each expert independently acquires knowledge within its specialized domain, while a gating network dynamically controls the contribution of each expert.
However, in traditional MoE systems, experts are initialized randomly. This paper aims to specialize each expert in learning the personalized information of individual vehicles. To achieve this, we propose a Dynamic Mixture-of-Experts (DMoE) module.
It leverages the MoE principle to assign each agent a dynamically generated expert. These experts are conditioned on the local features of each agent and capture distinct, agent-specific patterns. 
All expert outputs are then adaptively fused through a gated aggregation mechanism that accounts for reliability and complementarity. 
This formulation allows our framework to go beyond uniform fusion, enabling it to retain unique perceptual cues from different viewpoints while still integrating shared semantics. 
To further encourage expert diversity and prevent representational collapse, we introduce a Dynamic Expert Metric Loss (DEML), which promotes proximity between each expert and the fused feature while pushing different experts apart. 

We verify the effectiveness of our framework on two challenging collaborative perception benchmarks. CoBEVMoE achieves notable gains across both camera and LiDAR modalities, with up to \textbf{${+}1.5\%$ IoU} improvement on BEV segmentation and \textbf{${+}3.0\%$ AP@0.5} gain on 3D detection, outperforming strong baselines such as CoBEVT~\cite{xu2022cobevt} and AttFuse~\cite{xu2022opencood}. 
These results demonstrate the advantages of our expert-based heterogeneous fusion design.

Our main contributions are summarized as follows:

\begin{itemize}
    \item We rethink the precise fusion of multi-source information in collaborative perception from the perspective of agent-specific diversity and propose a novel dynamic mixture-of-experts-based framework, CoBEVMoE, to improve collaborative perception accuracy.

    \item We design an expert kernel dynamic initialization Dynamic Mixture-of-Experts module, which guides the learning direction of experts using the personalized features of agents. Simultaneously, a Dynamic Expert Metric Loss is designed to ensure the differentiation among experts, thereby guaranteeing the effective learning of personalized features.

    \item We verify the proposed CoBEVMoE method on two large-scale collaborative perception datasets. CoBEVMoE consistently outperforms strong baselines in 3D object detection and semantic segmentation.
\end{itemize}

\section{Related Work}
\begin{figure*}[h]
    \centering
    \includegraphics[width=0.85\linewidth]{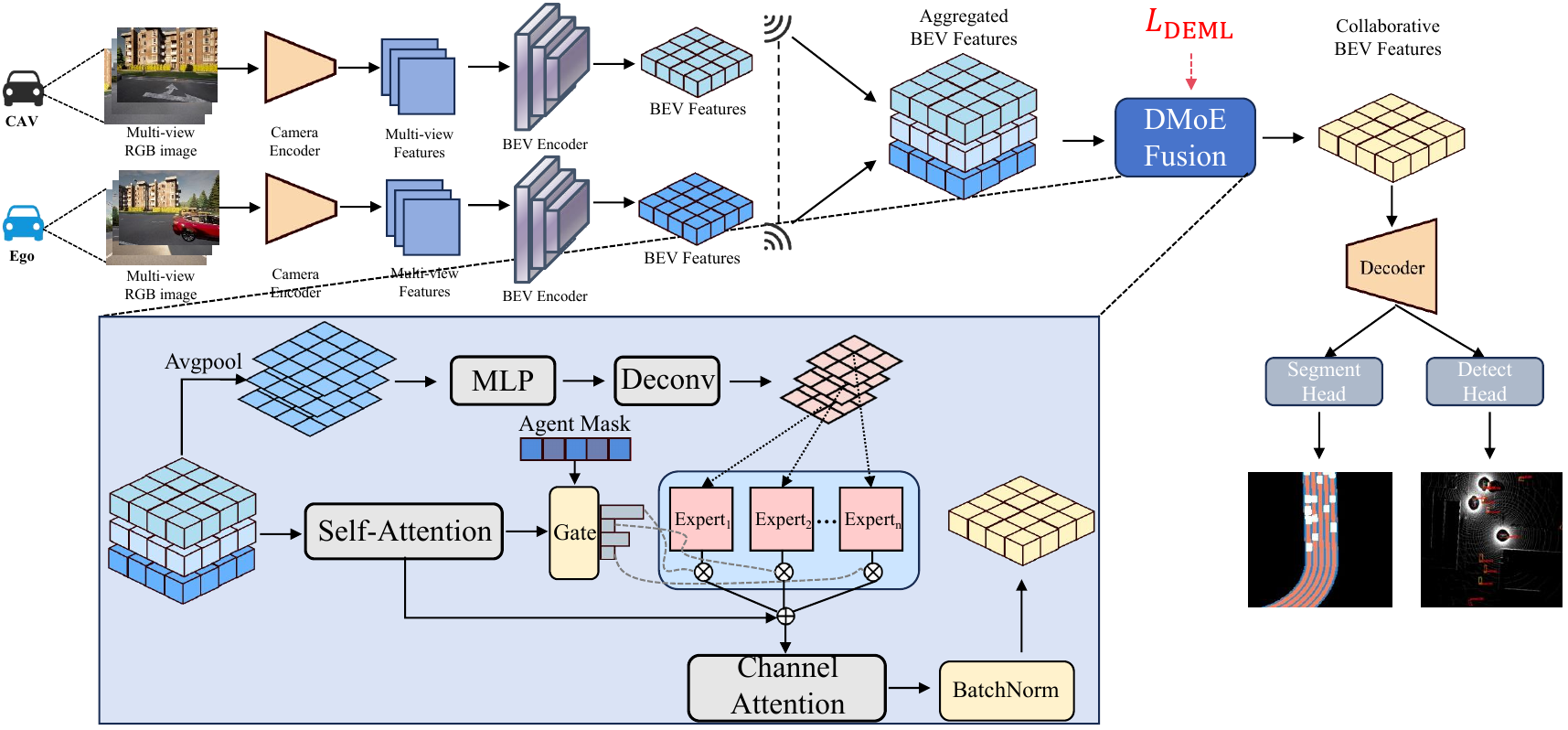}
    \vskip-1ex
    \caption{Illustration of the proposed \textbf{CoBEVMoE} framework. Each agent first extracts BEV features from its raw sensory inputs, which are then spatially aligned and transmitted to a central aggregator. A Dynamic Mixture-of-Experts (DMoE) module generates agent-specific expert kernels and performs adaptive feature aggregation through a learned gating mechanism. The resulting fused representation is subsequently decoded for downstream perception tasks, including semantic segmentation and 3D object detection. To encourage inter-expert diversity while maintaining consistency with the fused feature, we introduce a Dynamic Expert Metric Loss (DEML), depicted in $\mathcal{L}_\text{DEML}$ in the figure.  This loss promotes diverse yet coherent expert representations, ultimately enhancing the quality of the final fusion.}
    \label{fig:coBEVmoe}
    \vskip-3ex
\end{figure*}

\subsection{Collaborative Perception}

Most autonomous driving perception systems are built upon single-agent sensor data. 
While multi-sensor fusion improves perception accuracy, it still suffers from blind spots and a limited sensing range.
Collaborative perception addresses these issues by enabling information sharing among multiple agents, leading to more comprehensive environmental awareness~\cite{han2023collaborative,9676458,10801337,10802601}.
According to the fusion stage, collaborative perception methods are generally categorized into early, intermediate, and late fusion.
Early fusion aggregates raw sensor data~\cite{chen2019cooper}, but incurs prohibitive communication costs.
Late fusion merges individual perception outputs~\cite{dair-v2x}, which is often sensitive to noise and error accumulation.
Intermediate fusion performs collaboration at the feature level, offering a favorable trade-off between performance and communication efficiency, and has therefore become the dominant paradigm in recent studies~\cite{chen2019f,wang2020v2vnet,lu2024an,gao2025stamp}.
V2VNet~\cite{wang2020v2vnet} and DiscoNet~\cite{li2021learning} introduced intermediate fusion frameworks where BEV features were fused through a graph neural network. 
Recently, HEAL~\cite{lu2024an} and STAMP~\cite{gao2025stamp} addressed scalability and heterogeneity by pruning or learning dynamic communication graphs.

Subsequent research explored various feature fusion mechanisms to better handle spatial misalignment and inter-agent redundancy.
AttFusion~\cite{xu2022opencood} and V2X-ViT~\cite{xu2022v2xvit} incorporated attention-based modules to weigh features from different agents adaptively. 
CoBEVT~\cite{xu2022cobevt} proposed a Swin Transformer-based fusion backbone with window-grid alternating attention, enhancing local-global spatial interactions. 
Additionally, other works focused on robustness issues such as communication delays~\cite{lu2023robust, wei2023robust, lei2022syncnet, 11077784, Zhang_2024_CVPR}, positioning errors~\cite{9682601, pmlr-v155-vadivelu21a }, or adverse weather conditions~\cite{li2024v2x,zhang2024dsrc,ma2023macp} that may occur during the collaborative perception process. 
In practice, agents may observe the same scene from distinct perspectives, yielding non-overlapping but informative cues. 
Existing fusion strategies~\cite{wang2020v2vnet,xu2022cobevt} often suppress these subtle differences through averaging or global attention, which may degrade the final representation in complex, occluded scenes. 
To address this, our method introduces dynamically generated expert kernels that are initialized from unique features of each agent, allowing the model to extract and preserve unique perceptual cues during the fusion process.

\subsection{Mixture-of-Experts Models}

Mixture-of-Experts (MoE)~\cite{shazeer2017outrageously} is a conditional computation architecture that assigns input samples to a sparse subset of specialized expert networks via a gating mechanism. This approach has proven effective in scaling deep networks while maintaining diversity and efficiency. In NLP, models such as GShard~\cite{lepikhin2020gshard} and Switch Transformer~\cite{fedus2022switch} leverage MoE to distribute computation and capture linguistic variation. In vision, MoE has been applied to tasks like image classification~\cite{riquelme2021scaling} and video understanding~\cite{yang2022mvitv2moe}, showing strong performance gains on heterogeneous data.

Recent studies have extended MoE to multi-modal and multi-view perception. MetaBEV~\cite{ge2023metaBEV} incorporated an attention-based mixture of experts mechanism to learn modality-specific weights, enabling the adaptive fusion of reliable multi-modal features.
MapExpert~\cite{zhang2024mapexpertonlinehdmap} used MoE to decode various non-cubic map elements, improving map accuracy and efficiency. 
MoE-Fusion~\cite{Cao_2023_ICCV} applied MoE to infrared and visible image fusion, achieving superior results. 

These approaches directly apply the classical MoE paradigm to perception tasks, where each expert is learned implicitly within the model without a predefined or specialized learning objective.
In contrast, this paper aims to anchor the learning objectives of the experts. To this end, we build upon the MoE foundation and propose a dynamic expert kernel, which leverages features from different agents to guide and specialize the learning direction of each expert.

\section{Methodology}

\subsection{Problem Formulation}

Let $\mathcal{A} = \{1, 2, \dots, N\}$ denote a set of $N$ collaborating agents. Each agent $i \in \mathcal{A}$ is equipped with multiple cameras (or other sensors) to observe its surroundings and generate local multi-view image features. These features are transformed into a bird's-eye-view (BEV) representation using a shared image-to-BEV encoder, yielding a local BEV feature map $\mathbf{x}_i \in \mathbb{R}^{C \times H \times W}$, where $C$ is the feature dimension and $H, W$ are spatial dimensions.

Each agent then transmits its BEV features to a central fusion module (or to neighboring agents), which aligns the features via spatial transformation using ego-pose information. The aligned features from all agents form a set:
\begin{equation}
\mathcal{X} = \{ \mathbf{x}_1, \mathbf{x}_2, \dots, \mathbf{x}_N \}, \quad \mathbf{x}_i \in \mathbb{R}^{C \times H \times W}.
\end{equation}

The goal of the collaborative perception system is to fuse the set $\mathcal{X}$ into a unified representation $\mathbf{F}_{\text{fused}} \in \mathbb{R}^{C \times H \times W}$ that accurately describes the shared environment. This fused representation is then used for downstream tasks such as 3D object detection or BEV semantic segmentation.

The key challenge lies in effectively leveraging both the \textit{redundancy} (\textit{i.e.}, overlapping observations of the same object) and the \textit{heterogeneity} (\textit{i.e.}, complementary views and occlusion differences) among agents. To address this, we adopt an \textit{intermediate fusion} paradigm, which provides a balanced compromise between early fusion, where raw data are directly shared, and late fusion, where only decision-level information is exchanged, by transmitting intermediate features among agents. 

\subsection{Overview}
We revisit the collaborative perception problem from the perspective of agent-specific diversity and propose a new framework, CoBEVMoE. The framework consists of an image encoder, a BEV encoder, an intermediate collaboration module, a decoder, and a task head. Specifically, the image encoder (\textit{i.e.}, ResNet34) extracts multi-view features from raw images, and the BEV encoder (\textit{i.e.}, SinBEV Transformer) transforms them into BEV features. The intermediate collaboration module integrates our proposed Dynamic Mixture of Experts (DMoE) with Dynamic Expert Metric Loss (DEML) to fuse BEV features from all agents into a unified representation that improves perception. The decoder then processes the fused features, while the task head produces either BEV semantic segmentation or 3D object detection results. 
Our focus is on leveraging the diversity of agent perception features during collaboration to improve accuracy. To this end, we design the DMoE module with a DEML optimisation constraint.

\subsection{Dynamic Mixture-of-Experts Fusion}

Although prior work has demonstrated that MoE architecture can enhance the integration of multi-source features, conventional MoE architectures are ill-suited for multi-agent collaborative perception due to the divergent observational perspectives of each agent. To tackle the challenge of fusing heterogeneous features in multi-agent collaboration, we propose a novel intermediate collaboration module dubbed DMoE. It is designed to explicitly model both inter-agent relationships and agent-specific feature variations, thereby further advancing performance. Motivated by these limitations, we design a Dynamic Mixture of Experts (DMoE) module tailored for multi-agent collaboration, which adaptively learns and emphasizes the key contributions from each agent and explicitly models both the shared and agent-specific features in collaborative perception.

\paragraph{Expert Kernel Generation}

We first describe the construction of each expert in the proposed DMoE module. Unlike conventional designs where parameters are solely optimized via gradient descent, the parameters of each expert are dynamically generated conditioned on the input BEV features. Specifically, following the principle of dynamic convolution, each expert is instantiated as a dynamic convolutional layer whose kernels are adaptively generated from the input. After applying spatial transformation and feature aggregation across agents, we obtain the aligned BEV representations from the $N$ collaborating agents, denoted as:
\begin{equation}
\mathbf{X} = \big\{ \mathbf{x}_1, \mathbf{x}_2, \dots, \mathbf{x}_N \big\} \in \mathbb{R}^{B \times N \times C \times H \times W},
\end{equation}
where $B$ is the batch size, $N$ is the number of agents, and $C$, $H$, $W$ are the channel and spatial dimensions.

For each agent $k$, the BEV feature $\mathbf{x}_k$ is first subjected to spatial global average pooling:
\begin{equation}
\mathbf{z}_k = \text{AvgPool}(\mathbf{x}_k) \in \mathbb{R}^{B \times C' },
\end{equation}
where typically $C' < C \cdot H \cdot W$.

Then, the pooled vector $\mathbf{z}_k$ is processed by an MLP:
\begin{equation}
\mathbf{h}_k = \text{MLP}(\mathbf{z}_k) \in \mathbb{R}^{B \times d},
\end{equation}
and then decoded via a deconvolution block to generate a convolutional kernel:
\begin{equation}
\mathbf{W}_k = \text{Deconv}(\mathbf{h}_k) 
\in \mathbb{R}^{B \times C \times C \times 3 \times 3}.
\end{equation}

\paragraph{Gating Mechanism}
After the expert kernels are generated, their contributions to the output are determined through a gating unit. The gating unit is designed to generate gating functions based on the input features and serves as a fundamental component of the proposed DMoE. It is responsible for coordinating the participation of each expert and combining their outputs. Specifically, we apply global average pooling along the channel dimension of the input collaborative features $\mathbf{F_c} \in \mathbb{R}^{B \times C \times H \times W}$, and then use a softmax function to generate the weights $\boldsymbol{\alpha}$ for each expert:
\begin{equation}
\boldsymbol{\alpha} = \text{softmax}\big( \text{Gate}(\text{AvgPool}(\mathbf{F}_{c})) \big) \in \mathbb{R}^{B \times N}.
\end{equation}

To handle invalid agents, we apply a binary mask $\mathbf{M} \in \{0,1\}^{B \times L}$ to the gating weights:
\begin{equation}
\tilde{\boldsymbol{\alpha}} = \text{softmax}\big( \boldsymbol{\alpha} \odot \mathbf{M} + (1-\mathbf{M}) \cdot (-\infty) \big),
\end{equation}
where $\odot$ denotes element-wise multiplication.

\paragraph{Domain Expert Construction}

In the proposed DMoE module, we construct an expert for each agent, enabling it to extract key features from its unique perspective and organically fuse them for higher-quality collaborative perception. Each expert is designed as a dynamic convolutional layer. When it receives a feature input $\mathbf{F_c} \in \mathbb{R}^{B \times C \times H \times W}$, it utilizes the unique convolution kernel generated by the Expert Kernel Generation module to produce the corresponding feature $\mathbf{E}_k$:
\begin{equation}
\mathbf{E}_k = \mathbf{W}_k * \mathbf{F_c}.
\end{equation} 

Therefore, each expert acquires unique knowledge and contributes distinctly to the feature fusion.

\paragraph{DMoE Fusion Module}
The DMoE fusion module is composed of an expert kernel generation unit, a self-attention fusion layer, a gating mechanism, and a set of experts. Specifically, the aligned BEV features $\mathbf{X}$ are fed into the expert kernel generation module, and the generated expert kernels are assigned to each expert. Then, the $\mathbf{X}$ is directly input into the self-attention fusion layer to produce the preliminary fused BEV feature:
\begin{equation}
\mathbf{F}_{c} = \text{SelfAttn}(\mathbf{X}) \in \mathbb{R}^{B \times C \times H \times W}.
\end{equation}
Here, we use the original SwapFusion module from CoBEVT~\cite{xu2022cobevt} as our baseline.
Each expert applies its generated convolution kernel $\mathbf{W}_k$ to the preliminary fused feature $\mathbf{F}_c$ to produce the expert output $\mathbf{E}_k$.
All expert output features are stacked as:
\begin{equation}
\mathbf{E} = [\mathbf{E}_1, \mathbf{E}_2, \ldots, \mathbf{E}_N] \in \mathbb{R}^{B \times N \times C \times H \times W}.
\end{equation}

Finally, the expert weights obtained from the gating unit are combined with a residual connection to perform a weighted summation over the experts’ outputs, enhancing the stability and expressiveness of the MoE output feature. This residual structure preserves both the shared semantic information and the expert-specific enhancements in the fused representation. The overall process proceeds as follows:
\begin{equation}
\mathbf{F}_{\text{MoE}} = \sum_{k=1}^{L} \tilde{\alpha}_k \cdot \mathbf{E}_k.
\end{equation}
\begin{equation}
\mathbf{F}_{\text{res}} = \mathbf{F}_{c} + \mathbf{F}_{\text{MoE}}.
\end{equation}
 Unlike conventional fusion mechanisms~\cite{ge2023metaBEV,Cao_2023_ICCV}, the proposed DMoE not only captures spatial-temporal relationships across agents but also preserves the diversity of individual perspectives, enabling a more comprehensive and discriminative feature representation.

\subsection{Dynamic Expert Metric Loss}

Although the DMoE module effectively captures heterogeneous agent features, the extracted features may still lack sufficient regularization, potentially resulting in suboptimal performance. 
To address this, we introduce a Dynamic Expert Metric Loss (DEML) that serves as an internal regularizer for the DMoE module, aiming to prevent expert collapse by promoting functional diversity among agent-conditioned experts while maintaining consistency with the fused feature.
It differs from contrastive learning in that DEML operates on intra-sample expert outputs rather than learning instance-level representations.

\paragraph{Expert Triplet Construction}

Contrastive learning-based metric learning has shown strong performance in tasks that require fine-grained feature representations, as it directly optimizes the embedding space to preserve clear distinctions among embeddings. Motivated by this, we construct a unique triplet for each expert. 

The initial fused feature $\mathbf{F}_{c}$ output by the self-attention fusion layer contains relatively complete semantic information. 
We take $\mathbf{F}_{c}$ as the anchor to ensure that the experts' outputs do not deviate from the overall task objective. 
For each expert, we select its own output $\mathbf{E}_k$ as the positive sample, 
and choose the output of the most similar expert (based on mean squared error) from the remaining experts as the hardest negative sample $\mathbf{E}_j$
This ensures that the distance between positive and negative samples is sufficiently large, encouraging diversity among experts while learning the specific characteristics of each agent's features:
\begin{equation}
{T}_{k}=(\mathbf{F_c}, \mathbf{E_k}, \mathbf{E_j}),
\end{equation}
\begin{equation}
    \mathbf{E_j} = \argmin_{i \neq k} \text{MSE}(\mathbf{E}_i, \mathbf{E}_k).
\end{equation}

\paragraph{Expert Metric Loss}
Based on the constructed triplets, we define an expert-specific loss to guide the optimization process. Specifically, the $L_2$ distance is employed to measure the distances between the anchor, positive, and negative samples. The objective is to minimize the distance between the anchor and positive sample while ensuring that the distance between the positive and negative samples exceeds a predefined margin. This encourages the positive sample to remain close to the anchor and the negative sample to stay farther away. The triplet loss for expert $k$ is then defined as follows:
\begin{align}
d_\text{pos}^k &= \frac{1}{D} \| \mathbf{E}_k - \mathbf{F_c} \|_2^2, \\
d_\text{neg}^k &= \min_{j \neq k} \frac{1}{D} \| \mathbf{E}_k - \mathbf{E}_j \|_2^2.
\end{align}
\begin{equation}
\mathcal{L}_{\text{triplet}}^k =
\max\big( 0, d_\text{pos}^k - d_\text{neg}^k + m \big),
\end{equation}
where $m{>}0$ is a margin hyperparameter. And the overall expert regularization loss is formulated as:
\begin{equation}
\mathcal{L}_{\text{DEML}} =
\frac{1}{N_{\text{valid}}} \sum_k \big( d_\text{pos}^k + \beta \cdot \mathcal{L}_{\text{triplet}}^k \big),
\end{equation}
where $N_{\text{valid}}$ represents the number of valid experts, and $\beta$ is a balancing coefficient that weights the two terms.

Finally, we integrate the expert-specific DEML, scaled by a balancing hyperparameter $\lambda$, with the original loss objective to obtain the total training loss:
\begin{equation}
\mathcal{L}_{\text{total}} = \mathcal{L}_{\text{task}} + \lambda \cdot \mathcal{L}_{\text{DEML}},
\end{equation}
where $\mathcal{L}_{\text{task}}$ denotes the task-specific loss corresponding to different tasks, such as segmentation or detection.

\section{Experiments}

\subsection{Datasets}
To evaluate the effectiveness of the proposed method, we conducted extensive experiments on both the simulated dataset OPV2V~\cite{xu2022opencood} and the real-world dataset DAIR-V2X-C~\cite{dair-v2x} for comprehensive verification. 

\textbf{OPV2V} is a large-scale vehicle-to-vehicle collaborative perception dataset collected in CARLA and the cooperative driving automation tool OpenCDA. It comprises $12,000$ LiDAR point cloud frames and RGB images with $230,000$ annotations of 3D bounding boxes, with an image resolution of $600{\times}800$, providing a diverse set of scenarios for evaluating perception algorithms.

\textbf{DAIR-V2X-C} is a widely-used real-world vehicle-to-infrastructure collaborative perception dataset, consisting of $38,845$ frames of image data and point cloud data, with an image resolution of $1080{\times}1920$, covering diverse urban driving scenarios for evaluation.

\subsection{Evaluation Metrics}
To ensure fair and comprehensive evaluation, we adopt task-specific metrics for both BEV segmentation and 3D object detection. 

For the \textbf{BEV semantic segmentation} task on OPV2V, we report the Intersection over Union (IoU) for each semantic category, including \emph{vehicle}, \emph{drivable area}, and \emph{lane}. The IoU metric is defined as the ratio of the intersection to the union between predicted and ground truth masks, which effectively captures the overlap quality of semantic regions. Higher IoU indicates better segmentation accuracy and robustness against occlusion and viewpoint changes. 

For the \textbf{3D object detection} task on DAIR-V2X-C, we employ the Average Precision (AP) at different Intersection-over-Union thresholds, specifically AP@0.3 and AP@0.5, following standard practice in V2X perception. The AP metric measures the area under the precision-recall curve, reflecting the trade-off between detection precision and recall. The use of multiple IoU thresholds allows us to evaluate both coarse and fine-grained localization accuracy of detected objects in collaborative settings.

To quantify inter-expert diversity, we report \emph{Pairwise Cosine Diversity (PCD)}.
Given expert outputs $\{E_k\}_{k=1}^{N_\text{valid}}$, we obtain compact vectors by global average pooling:
$\mathbf{v}_k=\mathrm{GAP}(E_k)\in\mathbb{R}^{B\times C}$.
We compute the average pairwise cosine similarity
\begin{equation}
\mathrm{ES}=\frac{2}{N_\text{valid}(N_\text{valid}-1)}\sum_{i<j}\mathbb{E}_b\left[
\frac{\mathbf{v}_i^{(b)}\cdot \mathbf{v}_j^{(b)}}{\|\mathbf{v}_i^{(b)}\|_2\|\mathbf{v}_j^{(b)}\|_2+\epsilon}
\right],
\end{equation}
and define $\mathrm{PCD}=1-\mathrm{ES}$, where higher PCD indicates stronger diversity among experts.

\begin{table}[t]
    \centering
    \small
    \caption{IOU of BEV segmentation on OPV2V dataset camera-track.}
    \vskip-1ex
	\resizebox{0.48\textwidth}{!}{
		\setlength\tabcolsep{8pt}
		\renewcommand\arraystretch{1.0}
		\begin{tabular}{c||c|c|c}
			\hline
			\rowcolor{mygray}
        Methods & Veh. & Dr.Area & Lane  \\ \hline\hline
         F-Cooper\cite{chen2019f} & 52.5 & 59.0 & 47.2 \\
         AttFuse\cite{xu2022opencood} & 51.9 & 60.5 & 47.9 \\
         DiscoNet\cite{li2021learning} & 52.9 & 57.6 & 43.4 \\
         V2VNet\cite{wang2020v2vnet} & 53.5 & 58.2 & 43.8 \\ 
         CoBEVT\cite{xu2022cobevt} &  60.4 & 61.2 & 48.8 \\ \hline
         Ours (CoBEVMoE) &  \textbf{61.9} \color{mygreen}{(\textbf{{+1.5}})}& \textbf{61.8} \color{mygreen}{(\textbf{{+0.6}})} & \textbf{50.5} \color{mygreen}{(\textbf{{+1.7}})} \\ \hline
    \end{tabular}
    }
    
    \label{tab:opv2v}
    \vskip-3ex
\end{table}

\subsection{Experimental Setup}
\textbf{Implementation Details.} 
Following CoBEVT~\cite{xu2022cobevt}, we assume all the AVs have a $70m$ communication range, and all the vehicles outside this broadcasting radius of the ego vehicle will not have any collaboration. 
For the OPV2V camera-track BEV semantic segmentation, we choose ResNet34 as the image feature extractor. The transmitted BEV intermediate representation has a resolution of $32{\times}32{\times} 128$. 
For the DAIR-V2X LiDAR-track 3D object detection, we select PointPillar~\cite{lang2019pointpillars} as the point cloud feature extractor and set the voxel resolution as $(0.4, 0.4, 4)$ on the x, y, and z-axis. The architecture settings are the same as \cite{lang2019pointpillars}. 
The MLP used in the dynamic kernel generator consists of two fully connected layers with a hidden dimension of $128$ and ReLU activation.
The kernel generator employs a transposed convolution with kernel size $3\times3$, stride $1$, and no padding, which maps a $128$-dimensional latent embedding to dynamic convolutional kernels of size $128\times128\times3\times3$.
For the Dynamic Expert Metric Loss (DEML), we set the margin $m=0.5$ and the weighting factor $\beta=1.0$ for all experiments.
We train the whole model end-to-end with the Adam optimizer and the cosine annealing learning rate scheduler. We train all models on a single NVIDIA RTX A6000 GPU with $150$ epochs and a batch size of $2$.

\subsection{Quantitative and Qualitative Evaluation}

\begin{figure}[h]
  \centering
  \includegraphics[width=\linewidth]{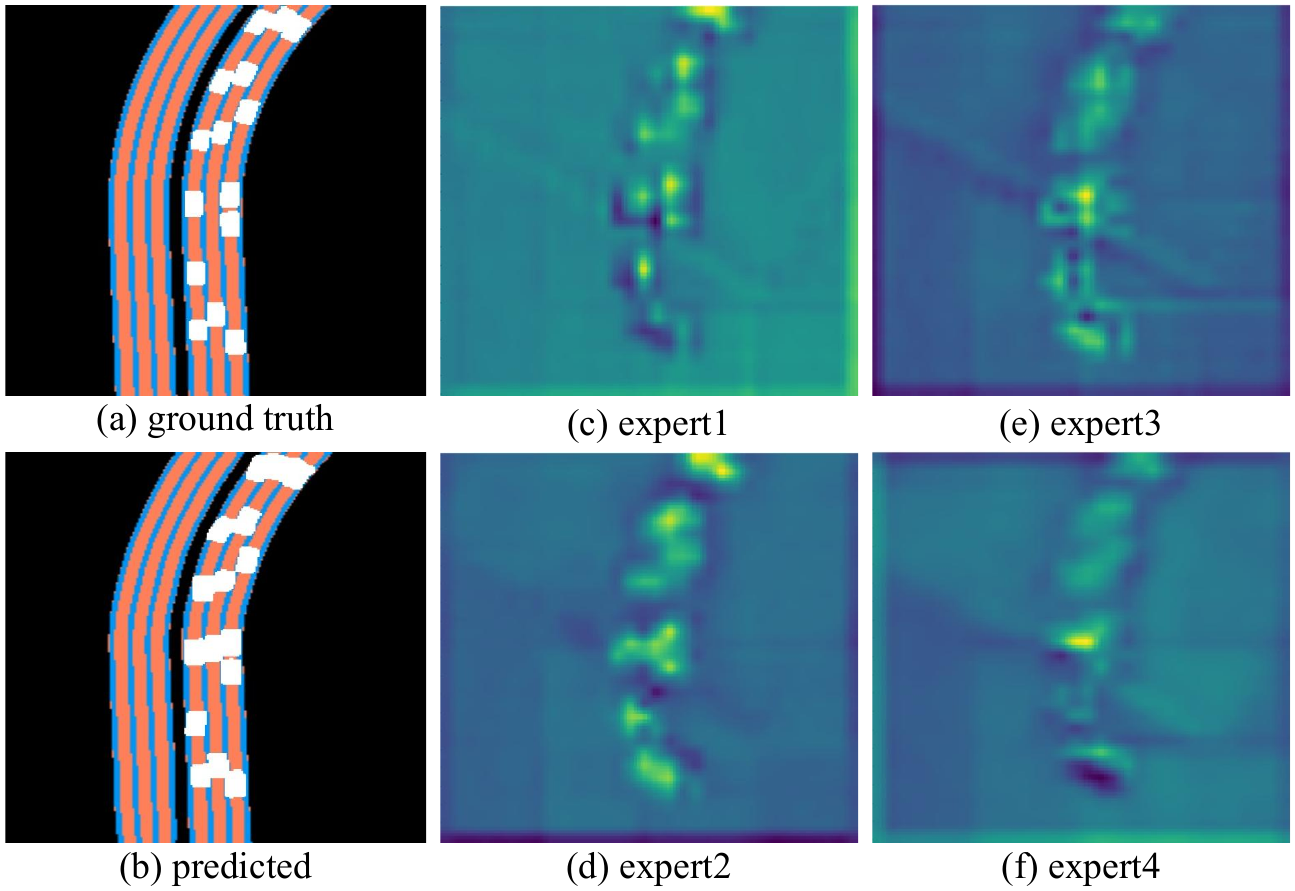}
  \vskip-1ex
  \caption{Qualitative comparison among different expert feature maps. (a) shows the ground truth BEV semantic map. (b) shows the predicted BEV semantic map, while (c-f) depict the activation maps of four different experts in our DMoE module. The four experts focus on different spatial patterns, indicating the diversity in feature extraction. This complementary behavior contributes to a more comprehensive and robust feature fusion in the downstream task.
  }
  \label{fig:experts_compare_1}
  \vskip-3ex
\end{figure}

\textbf{OPV2V camera-track results.}  
Table~\ref{tab:opv2v} reports the performance of our method and several strong baselines on the camera-based BEV segmentation task. Compared to existing methods such as AttFuse~\cite{xu2022opencood} and CoBEVT~\cite{xu2022cobevt}, our proposed CoBEVMoE consistently achieves higher IoU across all categories. Compared to CoBEVT, our method improves the IoU for vehicle, drivable area, and lane segmentation by ${+}1.5\%$, ${+}0.6\%$, and ${+}1.7\%$, respectively. 
This performance gain can be attributed to the joint effect of our DMoE module and the DEML. The DMoE module enables each agent to dynamically capture both redundant and complementary information, while the DEML encourages feature diversity across agents, mitigating information collapse. As shown in Fig.~\ref{fig:experts_compare_1}, different experts focus on diverse spatial patterns, complementing each other and contributing to a more comprehensive fusion representation.

\textbf{DAIR-V2X LiDAR-track results.}
We further evaluate the 3D object detection performance of CoBEVMoE on the real-world DAIR-V2X-C dataset, which provides LiDAR point clouds under an infrastructure-to-vehicle collaboration setting. 
As shown in Table~\ref{tab:dairv2x}, our method outperforms CoBEVT by ${+}0.7\%$ on AP@0.3 and ${+}3.0\%$ on AP@0.5.
These results confirm that our framework not only excels in camera-based BEV segmentation but also generalizes effectively to geometric perception tasks in realistic V2X environments. The improvements demonstrate that the proposed DMoE module can handle sensor noise, occlusion, and spatial misalignment, while the expert-based fusion strategy enhances adaptability to diverse real-world deployment scenarios.

\begin{table}[!t]
    \small
    \caption{Average Precision (AP) of 3D detection on the DAIR-V2X LiDAR-track.}
    \vskip-1ex
	\resizebox{0.48\textwidth}{!}{
		\setlength\tabcolsep{8pt}
		\renewcommand\arraystretch{1.0}
		\begin{tabular}{c||c|c}
            \hline
			\rowcolor{mygray}
         Methods & AP@0.3 &AP@0.5 \\ \hline\hline
         
         F-Cooper\cite{chen2019f} & 72.3 & 62.0 \\
         AttFuse\cite{xu2022opencood} & 73.8 & 67.3 \\
         DiscoNet\cite{li2021learning} & 74.6 & 68.5 \\
         V2XViT\cite{xu2022v2xvit} & 77.6 & 52.1 \\
         CoBEVT\cite{xu2022cobevt} & 77.9 & 67.3 \\ \hline
         Ours (CoBEVMoE) & \textbf{78.6} \color{mygreen}{(\textbf{{+0.7}})} & \textbf{70.3} \color{mygreen}{(\textbf{{+3.0}})} \\ \hline
    \end{tabular}
    }
    
    \label{tab:dairv2x}
    \vskip-3ex
\end{table}

\begin{figure*}
  \centering
  \includegraphics[width=\linewidth]{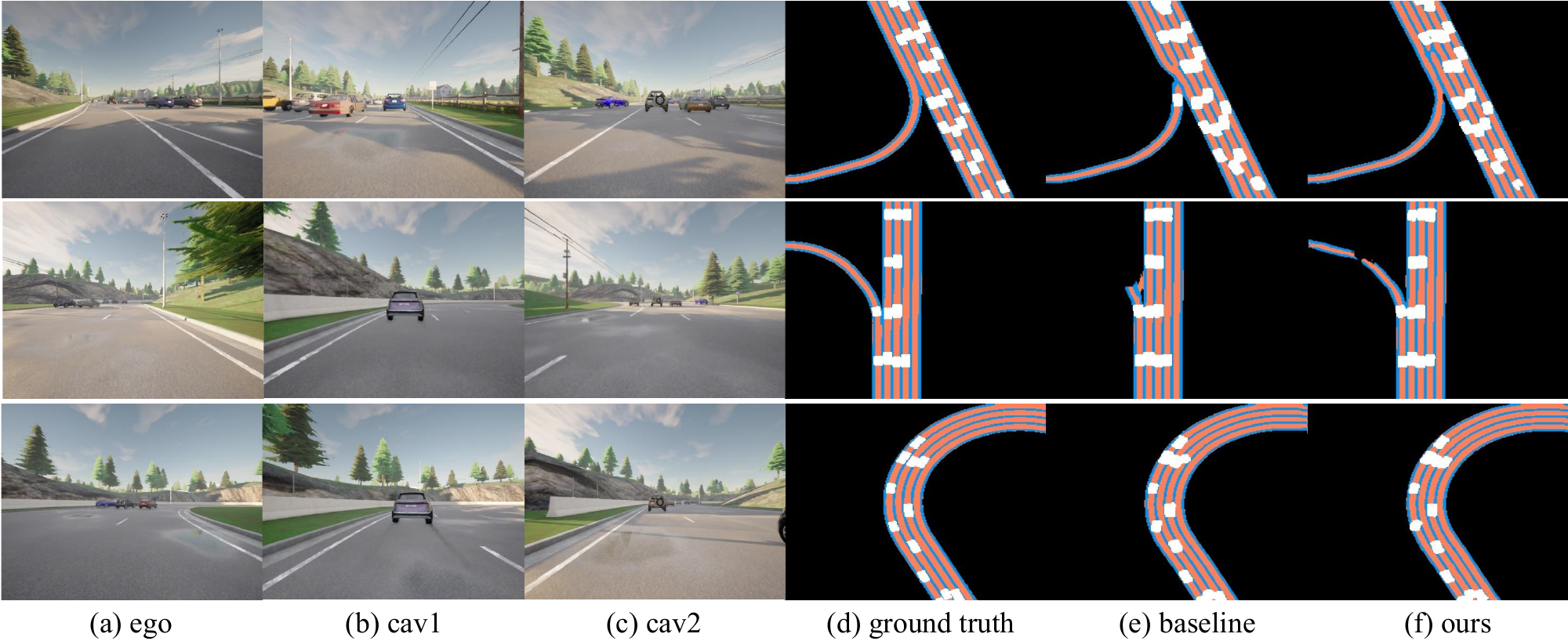}
  \vskip-1ex
  \caption{
  Qualitative results of collaborative BEV segmentation.
  From left to right: (a) ego vehicle's front camera image, (b–c) front camera images from two collaborating agents (cav1 and cav2), (d) ground truth BEV segmentation, (e) result of the baseline fusion method, and (f) result of our proposed \textbf{CoBEVMoE}.
  Compared to the baseline, our method produces more complete and precise semantic segmentation, especially for distant or partially occluded targets, demonstrating its ability to effectively integrate complementary views and retain agent-specific information.
  }
  \label{fig:qualitative_coBEVmoe}
  \vskip-3ex
\end{figure*}

Qualitative comparisons in Fig.~\ref{fig:qualitative_coBEVmoe} further demonstrate that CoBEVMoE generates more complete and accurate BEV segmentations than the baseline, particularly for distant or occluded targets.

\begin{figure}
  \centering
  \includegraphics[width=\linewidth]{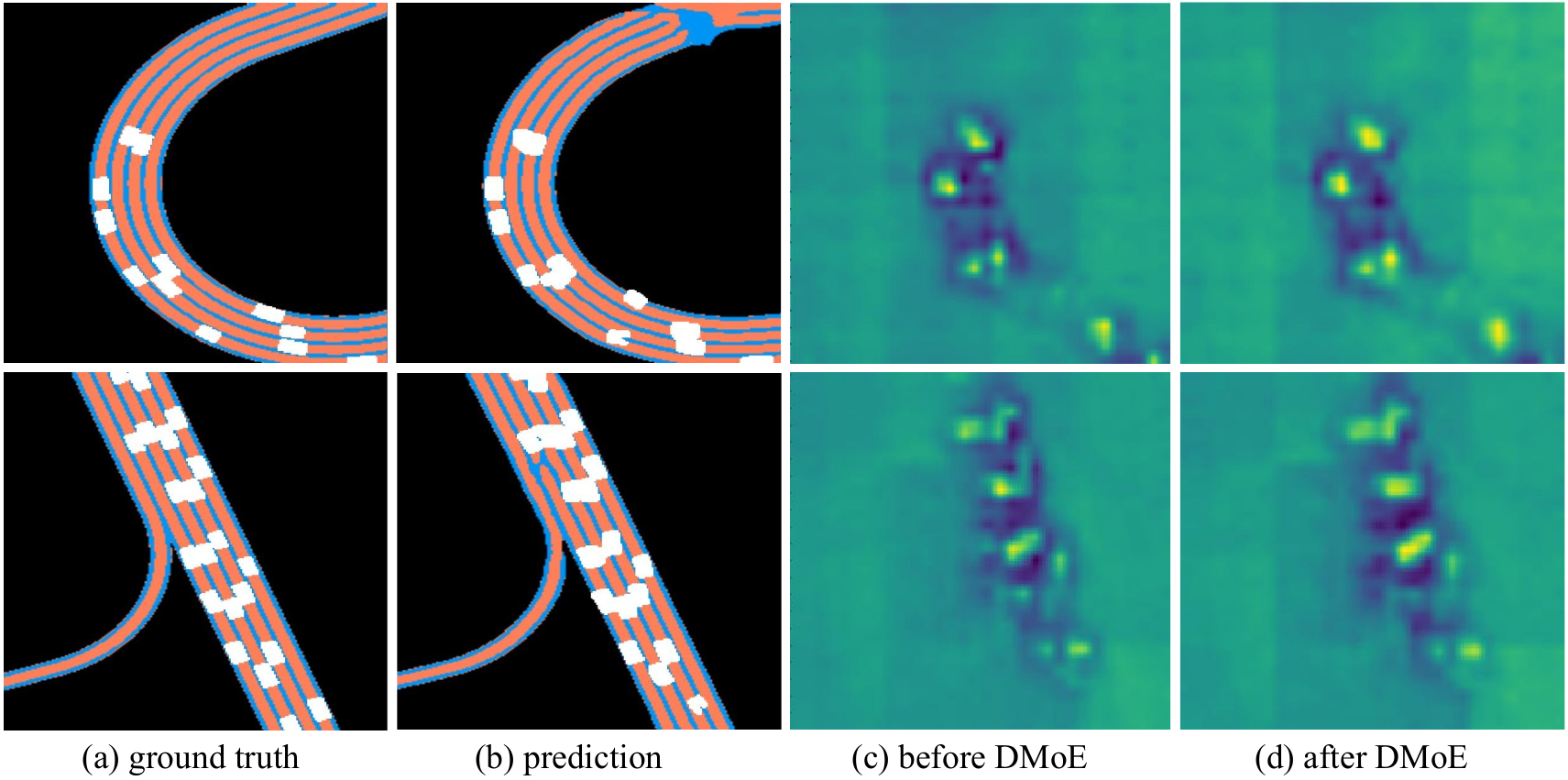}
  \vskip-1ex
  \caption{
  Visualization of BEV feature maps before and after the DMoE fusion module. (a) shows the ground truth BEV semantic map. (b) shows the predicted BEV semantic map. (c) presents the feature activation of a representative channel before DMoE fusion, while (d) displays the output after fusion. The DMoE-enhanced feature map demonstrates clearer and more focused activations, indicating that our expert-based fusion improves feature quality and semantic alignment.
}
  \label{fig:experts_compare_2}
  \vskip-3ex
\end{figure}

Moreover, Fig.~\ref{fig:experts_compare_2} visualizes BEV feature maps before and after DMoE fusion. The enhanced features exhibit clearer and more focused activations, indicating that our expert-based fusion improves feature quality and semantic alignment.

\subsection{Ablation Studies}

\textbf{Ablation studies on hyperparameters.} 
Table~\ref{tab:hyp} reports the ablation study on the loss weight $\lambda$ of the DEML, conducted on the OPV2V camera-based segmentation task. 
We vary $\lambda$ from $0$ to $1.0$ with an interval of $0.2$. As shown in the table, the performance remains relatively stable when $\lambda$ ranges from $0.0$ to $0.6$, with the best results achieved at $\lambda=0.4$, where vehicle, drivable area, and lane segmentation reach their highest IoUs. 
This indicates that a moderate loss weight provides sufficient regularization for the DMoE module, encouraging feature diversity without overwhelming the primary segmentation objective. 
In contrast, excessively small weights (\textit{e.g.}, $\lambda=0.2$) fail to impose effective constraints, while overly large weights (\textit{e.g.}, $\lambda \geq 0.8$) dominate the optimization, leading to degraded semantic segmentation performance. These results highlight the importance of balancing the expert regularization and the main task loss for stable and effective training.

\textbf{Component analysis.}  
Table~\ref{tab:ablation} presents an ablation study on the key components of CoBEVMoE. 
From the results, we can see that each proposed component contributes incremental improvements. Adding a Vanilla MoE improves IoU by ${+}0.4$ on \emph{vehicles}, while DMoE further boosts it by an additional ${+}0.5$. 
Incorporating the DEML yields a final gain of ${+}0.6$, leading to a total improvement of ${+}1.5$ in IoU over the CoBEVT baseline.
We further quantify expert diversity using PCD. As shown in Table \ref{tab:ablation}, enabling DEML significantly increases PCD from $0.10$ to $1.53$, which consistently correlates with the performance gains, indicating that DEML encourages agent-relevant expert diversity rather than arbitrary dispersion.
Here, the \emph{Vanilla MoE} baseline employs \textbf{five static experts}, which are \textbf{randomly initialized and jointly optimized during training}, while \textbf{sharing the same gating network as DMoE}.
We make the following observations:

\begin{table}[!t]
    \centering
    \small
    \caption{Ablation results of hyperparameters $\lambda$ on the OPV2V Camera-track.}
    \vskip-1ex
	\resizebox{0.5\textwidth}{!}{
		\setlength\tabcolsep{16pt}
		\renewcommand\arraystretch{1.0}
		\begin{tabular}{c||c|c|c}
			\hline
			\rowcolor{mygray}
        $\lambda$ & Veh. & Dr.Area & Lane  \\ \hline\hline
         0 & 61.3 & 61.6 & 50.1 \\
         0.2 & 61.4 & 61.8 & 50.2 \\
         0.4 & \textbf{61.9} & \textbf{61.8} & \textbf{50.5} \\
         0.6 & 61.4 & 61.7 & 50.2 \\ 
         0.8 & 60.7 & 61.3 & 49.2 \\
         1.0 & 59.8 &  61.0& 48.7  \\ \hline
    \end{tabular}
    }
    
    \label{tab:hyp}
    \vskip-3ex
\end{table}

\begin{itemize}
    \item \textbf{Vanilla MoE \textit{vs.} CoBEVT:} Simply using a standard mixture-of-experts module with static experts yields only marginal improvements. This indicates that without expert-specific design and regulation, the expert system fails to effectively disentangle useful agent-specific cues.
    
    \item \textbf{DMoE \textit{vs.} Vanilla MoE:} Introducing dynamic kernel generation for each expert significantly boosts performance, as the learned kernels are conditioned on each agent's local context. This dynamic adaptability enhances the model’s capacity to capture heterogeneity across agents.
    
    \item \textbf{DMoE + DEML:} The proposed Dynamic Expert Metric Loss further regularizes the expert outputs by explicitly encouraging diversity among experts while maintaining alignment with the fused feature. This alleviates feature redundancy and improves discriminability, leading to consistent gains across all segmentation categories.
\end{itemize}

Together, these results demonstrate that our proposed DMoE module and DEML are complementary and crucial for high-quality multi-agent feature fusion.

\begin{table}[!t]
\centering
\caption{Ablation study on the OPV2V dataset. ``DMoE'' denotes our Dynamic Mixture-of-Experts module. ``DEML'' indicates the use of the proposed Dynamic Expert Metric Loss.}
\vskip-1ex
\label{tab:ablation}
\begin{tabular}{lcccc}
\toprule
\textbf{Method} & Veh. & Dr.Area & Lane & PCD\\
\midrule
Baseline (CoBEVT) & 60.4 & 61.2 & 48.8 & --\\
+ Vanilla MoE       & 60.8 & 61.2 & 49.1 & 0.10\\
+ DMoE (Ours)       & 61.3 & 61.6 & 50.1 & 1.07\\
+ DMoE + DEML (Ours) & \textbf{61.9} & \textbf{61.8} & \textbf{50.5} & 1.53\\
\bottomrule
\end{tabular}
\vskip-3ex
\end{table}

\section{Conclusion}
In this paper, we propose CoBEVMoE, a collaborative perception framework that leverages a dynamic mixture-of-experts module and a dynamic
expert metric loss to capture both shared and agent-specific features during intermediate fusion. Experiments on OPV2V and DAIR-V2X demonstrate that our approach consistently outperforms strong baselines in semantic segmentation and 3D object detection, confirming its effectiveness across modalities and scenarios. In future work, we aim to extend CoBEVMoE to bandwidth-constrained settings and explore its integration with temporal modeling for multi-agent prediction and planning.

\bibliographystyle{IEEEtran}
\bibliography{bib.bib}

\end{document}